\documentclass[11pt]{article}
\usepackage{acl}
\usepackage{times}
\usepackage{latexsym}
\usepackage{graphicx}
\usepackage{booktabs}
\usepackage{amsmath}
\usepackage{multirow}
\usepackage{xcolor}
\usepackage{colortbl}
\usepackage{enumitem}  % For compact lists

\title{One Word Is Not Enough: Simple Prompts Improve Word Embeddings}

\author{Rajeev Ranjan \\
  GodelLabs \\
  \texttt{rajeev@godellabs.ai}}

\begin{document}
\maketitle

\begin{abstract}
Text embedding models are designed for sentence-level applications like retrieval and semantic similarity, and are primarily evaluated on sentence-level benchmarks. Their behavior on isolated words is less understood. We show that simply prepending semantic prompts to words before embedding substantially improves word similarity correlations. Testing 7 text embedding models, including text-embedding-3-large (OpenAI), embed-english-v3.0 (Cohere), voyage-3 (Voyage AI), all-mpnet-base-v2, and Qwen3-Embedding-8B, on 3 standard benchmarks (SimLex-999, WordSim-353, MEN-3000), we find that prompts like ``meaning: \{word\}'' or ``Represent the semantic concept: \{word\}'' improve Spearman correlations by up to +0.29 on SimLex-999. Some models fail completely on bare words ($\rho \approx 0$) but recover with prompts (+0.73 improvement). Our best results achieve $\rho=0.692$ on SimLex-999 with embed-english-v3.0 (Cohere), $\rho=0.811$ on WordSim-353, and $\rho=0.855$ on MEN-3000 with text-embedding-3-large (OpenAI). These results outperform classic static embeddings like Word2Vec ($\rho=0.40$) and even the best static method LexVec ($\rho=0.48$) on SimLex-999, establishing a new state-of-the-art for pure embedding methods. This zero-shot technique requires no training and works with any text embedding model.
\end{abstract}

\section{Introduction}

Word embeddings revolutionized NLP when Word2Vec \cite{mikolov2013word2vec} and GloVe \cite{pennington2014glove} showed that words could be represented as dense vectors capturing semantic relationships. Follow-up work like fastText \cite{bojanowski2017fasttext} and LexVec \cite{salle2016lexvec} improved performance through subword information and better matrix factorization. These static embeddings were trained explicitly on word co-occurrence patterns, producing high-quality word representations.

The field has since shifted to text embedding models optimized for retrieval and RAG applications. Open-source models like Sentence-BERT \cite{reimers2019sentence} use contrastive learning on datasets like SNLI \cite{bowman2015snli} and MNLI \cite{williams2018mnli}; commercial APIs (OpenAI, Cohere, Voyage) do not disclose their training procedures, though they are likely trained on sentence-level tasks. But what happens when we pass a single word like ``dog'' to these models?

This is potentially problematic for two reasons. First, isolated words may be \textbf{out-of-distribution}: if models primarily see full sentences during training, single-word inputs are unusual. Second, \textbf{tokenization may differ}: subword tokenizers often encode ``dog'' (sentence-initial) differently from `` dog'' (mid-sentence with space prefix). However, as we show, the effects of these factors are highly model-dependent.

Despite the shift to sentence-level models, word-level embeddings remain essential for many NLP applications. \textbf{Word sense disambiguation} relies on word representations to determine which meaning of a polysemous word (e.g., ``bank'') is intended in context, a prerequisite for machine translation and question answering. \textbf{Biomedical NLP} uses word embeddings to link clinical terms like ``heart attack'' to standardized ontologies (SNOMED-CT \cite{donnelly2006snomed}, MeSH \cite{lipscomb2000mesh}) for entity recognition and knowledge graph construction. \textbf{Retrieval systems} struggle with short queries: when a user searches for a single keyword, sentence embedding models trained on longer texts often fail to capture the query's semantics. Yet practitioners routinely use text embedding models for these word-level tasks without understanding how input formatting affects embedding quality.

We investigate a simple but effective technique: \textbf{prepending prompts to words before embedding}. For example, instead of embedding ``dog'' directly, we embed ``meaning: dog'' or ``Represent the semantic concept: dog.'' This modification yields large improvements on word similarity benchmarks: up to +0.29 correlation improvement for well-behaved models, and +0.73 for models that fail on bare words.

Our contributions are:
\begin{enumerate}
    \item We systematically evaluate how input formatting affects word embeddings from text embedding models, testing 8 conditions across 7 models.
    \item We discover that semantic prompts significantly improve word similarity correlations, with consistent gains across all tested models.
    \item We establish new state-of-the-art results for pure embedding methods on SimLex-999, WordSim-353, and MEN-3000.
\end{enumerate}

\section{Related Work}

\paragraph{Contextual Word Representations.}
\citet{ethayarajh2019contextual} showed that less than 5\% of variance in BERT's contextualized representations can be explained by static embeddings, motivating research on extracting word-level representations from contextual models. \citet{bommasani2020interpreting} proposed aggregating embeddings across multiple sentence contexts to create static word vectors: they sample sentences containing a target word from a corpus, compute contextualized embeddings in each context, and average them. While effective, this requires a large corpus and multiple forward passes per word. Our approach is different: instead of aggregating over many contexts, we provide a single short prompt (e.g., ``meaning: dog'') that acts as a lightweight pseudo-context, achieving similar benefits with a single forward pass and no corpus requirement.

\paragraph{Static Embeddings from Transformers.}
Recent work has developed efficient methods to extract static embeddings from transformers. Model2Vec \cite{tulkens2024model2vec} distills text embedding models into small static models by passing the entire vocabulary through a sentence transformer and applying PCA, achieving 500x speedup. WordLlama \cite{miller2024wordllama} extracts the token embedding layer from decoder LLMs (e.g., LLaMA) and trains a lightweight projection. Both methods create static lookup tables requiring distillation or training. Our approach is complementary: we use the full model at inference time with prompts, requiring no training and preserving the model's full representational capacity.

\paragraph{LLM-Based Word Embeddings.}
\citet{mahajan2024revisiting} evaluated word embeddings from billion-parameter LLMs (LLaMA2-7B, GPT ADA-002, PaLM) against classical methods. They found that smaller text embedding models (SBERT-style) nearly match LLMs on word-level tasks despite being 10-20x smaller, but both used bare words without any formatting. Our work shows that the gap between text embedding models and LLMs may partly stem from input formatting: with appropriate prompts, text embedding models achieve strong word similarity performance without the computational cost of LLMs. This suggests practitioners should consider prompting strategies before scaling to larger models.

\paragraph{Instruction-Tuned Embeddings.}
INSTRUCTOR \cite{su2023instructor} demonstrated that task-specific instructions improve sentence embeddings for various downstream tasks. Similarly, BGE \cite{xiao2023bge} and E5 \cite{wang2024improving} use query prefixes for retrieval tasks. However, these approaches focus on sentence-level tasks and require training with instructions. Our work shows that simple prompts work \textit{zero-shot} for word-level similarity without any training.

\paragraph{Tokenization Effects.}
\citet{garisoler2024impact} examined how subword splitting affects contextualized representations, finding that average pooling works best for multi-token words. However, no prior work has systematically evaluated input \textit{formatting} (e.g., space prefixes, semantic prompts) for word embeddings from text embedding models. We fill this gap with an evaluation across models and benchmarks.

\section{Method}

Given a word $w$, we create a prompted input $p(w)$ using a template function $p$. We then obtain an embedding $\mathbf{e} = f(p(w))$ where $f$ is a text embedding model. We evaluate embeddings by computing cosine similarity between word pairs and measuring Spearman correlation ($\rho$) with human similarity judgments.

\subsection{Prompt Conditions}

We test 8 conditions organized into two categories:

\paragraph{Formatting conditions} test tokenization effects:
\begin{itemize}[nosep,leftmargin=*]
    \item \textbf{bare}: $w$ (baseline)
    \item \textbf{leading\_space}: `` $w$'' (space prefix)
    \item \textbf{trailing\_space}: ``$w$ '' (space suffix)
    \item \textbf{both\_spaces}: `` $w$ '' (both)
\end{itemize}

\paragraph{Semantic conditions} test contextual priming:
\begin{itemize}[nosep,leftmargin=*]
    \item \textbf{the\_word}: ``the word $w$''
    \item \textbf{word\_colon}: ``word: $w$''
    \item \textbf{meaning\_colon}: ``meaning: $w$''
    \item \textbf{instruct\_semantic}: ``Represent the semantic concept: $w$''
\end{itemize}

The formatting conditions test whether whitespace affects embedding quality. We observe that some models are sensitive to whitespace (producing different embeddings for `` cat'' vs ``cat''), while others are insensitive (producing identical embeddings regardless of surrounding spaces). As shown in Tables~\ref{tab:full_simlex}--\ref{tab:full_men}, all-mpnet-base-v2, BGE-large-en-v1.5, and embed-english-v3.0 show identical scores across all space conditions, indicating whitespace insensitivity. In contrast, OpenAI, Voyage, and Qwen models show varying scores, indicating whitespace sensitivity. The semantic conditions test whether providing semantic context improves the embeddings.

\subsection{Models}

We evaluate 7 text embedding models representing both commercial APIs and open-source options. Commercial APIs include OpenAI's text-embedding-3-small and text-embedding-3-large \cite{openai2024embeddings}, Cohere's embed-english-v3.0 \cite{cohere2024embed}, and Voyage AI's voyage-3 \cite{voyageai2024embeddings}. Open-source models include all-mpnet-base-v2 \cite{song2020mpnet,reimers2019sentence}, BGE-large-en-v1.5 \cite{xiao2023bge}, and Qwen3-Embedding-8B \cite{qwen2024embedding}. Table~\ref{tab:models} summarizes the models, their embedding dimensions, and sources.

\subsection{Datasets}

We use three standard word similarity benchmarks:
\begin{itemize}[nosep,leftmargin=*]
    \item \textbf{SimLex-999} \cite{hill2015simlex}: 999 pairs rated for genuine semantic \textit{similarity} (not relatedness). This is the most challenging benchmark as it distinguishes similar words (``car''/``automobile'') from merely related ones (``car''/``road'').
    \item \textbf{WordSim-353} \cite{finkelstein2002placing}: 353 pairs measuring a mix of similarity and relatedness.
    \item \textbf{MEN-3000} \cite{bruni2012men}: 3000 pairs measuring semantic relatedness via crowdsourcing.
\end{itemize}

\begin{table}[h]
\centering
\small
\begin{tabular}{lll}
\toprule
\textbf{Model} & \textbf{Dim.} & \textbf{Source} \\
\midrule
\multicolumn{3}{c}{\textit{Commercial APIs}} \\
\midrule
text-embedding-3-small & 1536 & OpenAI \\
text-embedding-3-large & 3072 & OpenAI \\
embed-english-v3.0 & 1024 & Cohere \\
voyage-3 & 1024 & Voyage AI \\
\midrule
\multicolumn{3}{c}{\textit{Open-Source Models}} \\
\midrule
all-mpnet-base-v2 & 768 & Sentence-Trans. \\
BGE-large-en-v1.5 & 1024 & BAAI \\
Qwen3-Embedding-8B & 4096 & Alibaba \\
\bottomrule
\end{tabular}
\caption{Text embedding models evaluated with their embedding dimensions and sources.}
\label{tab:models}
\end{table}

\section{Results}

\begin{table}[t]
\centering
\small
\begin{tabular}{l|ccc}
\toprule
\textbf{Model} & \textbf{Bare} & \textbf{Best} & \textbf{$\Delta$} \\
\midrule
\multicolumn{4}{c}{\textit{SimLex-999}} \\
\midrule
embed-english-v3.0 & 0.600 & \textbf{0.692} & +0.092 \\
text-embed-3-small & 0.502 & 0.671 & +0.169 \\
BGE-large-en-v1.5 & 0.568 & 0.650 & +0.082 \\
text-embed-3-large & 0.566 & 0.654 & +0.088 \\
voyage-3 & -0.071 & 0.587 & +0.658 \\
Qwen3-Embed-8B & 0.286 & 0.567 & +0.281 \\
all-mpnet-base-v2 & 0.536 & 0.576 & +0.040 \\
\midrule
\multicolumn{4}{c}{\textit{WordSim-353}} \\
\midrule
text-embed-3-large & 0.723 & \textbf{0.811} & +0.088 \\
BGE-large-en-v1.5 & 0.723 & 0.805 & +0.082 \\
all-mpnet-base-v2 & 0.744 & 0.761 & +0.017 \\
embed-english-v3.0 & 0.715 & 0.758 & +0.043 \\
text-embed-3-small & 0.650 & 0.744 & +0.094 \\
voyage-3 & -0.078 & 0.715 & +0.793 \\
Qwen3-Embed-8B & 0.356 & 0.542 & +0.186 \\
\midrule
\multicolumn{4}{c}{\textit{MEN-3000}} \\
\midrule
text-embed-3-large & 0.784 & \textbf{0.855} & +0.071 \\
text-embed-3-small & 0.739 & 0.836 & +0.097 \\
voyage-3 & 0.096 & 0.826 & +0.730 \\
BGE-large-en-v1.5 & 0.738 & 0.819 & +0.081 \\
embed-english-v3.0 & 0.749 & 0.759 & +0.010 \\
all-mpnet-base-v2 & 0.774 & 0.805 & +0.031 \\
Qwen3-Embed-8B & 0.439 & 0.708 & +0.269 \\
\bottomrule
\end{tabular}
\caption{Word similarity results (Spearman $\rho$). Best prompted condition vs bare word baseline. Best overall results per dataset in bold. Model names abbreviated: text-embed = text-embedding, Embed = Embedding.}
\label{tab:main_results}
\end{table}

Table \ref{tab:main_results} shows that prompts improve word similarity correlations across all models and datasets. Key findings are summarized below.

\paragraph{Prompts provide substantial gains.} On SimLex-999, prompts improve correlations by +0.04 to +0.29 for models with reasonable baselines. Qwen3-Embedding-8B shows the largest improvement (+0.281), jumping from $\rho=0.286$ (\textbf{bare}) to $\rho=0.567$ with \textbf{instruct\_semantic}, nearly doubling its correlation. text-embedding-3-small improves from $\rho=0.502$ to $\rho=0.671$ (+0.169).

\paragraph{Some models fail completely on bare words.} voyage-3 produces near-zero or negative correlations on bare words ($\rho=-0.07$ on SimLex-999, $\rho=-0.08$ on WordSim-353, $\rho=0.10$ on MEN) but achieves competitive results with prompts ($\rho=0.587$, $0.715$, $0.826$ respectively). This suggests voyage-3 was trained primarily on sentences and cannot meaningfully process isolated tokens without context.

\paragraph{Best prompts vary by model.} For OpenAI's text-embedding-3 models, \textbf{instruct\_semantic} and \textbf{meaning\_colon} work best. For embed-english-v3.0, \textbf{instruct\_semantic} is optimal on SimLex-999 while \textbf{the\_word} performs best on WordSim-353. BGE-large-en-v1.5 responds best to \textbf{word\_colon} across all benchmarks. Qwen3-Embedding-8B benefits most from \textbf{instruct\_semantic} on SimLex-999 but from space-based formatting on WordSim-353 and MEN. This variation suggests models have different sensitivities to input formatting based on their training procedures.

\paragraph{Semantic prompts outperform formatting.} Across all models, semantic prompts consistently outperform simple formatting changes. For text-embedding-3-small, space prefixes provide modest gains (+0.02 on SimLex), while semantic prompts like \textbf{meaning\_colon} provide much larger improvements (+0.17 on SimLex). This suggests the benefit comes primarily from semantic priming rather than tokenization effects. Full results for all conditions are in Tables~\ref{tab:full_simlex}--\ref{tab:full_men}.

\subsection{State-of-the-Art Comparison}

\begin{table}[t]
\centering
\footnotesize
\setlength{\tabcolsep}{3pt}
\begin{tabular}{l|ccc|l}
\toprule
\textbf{Method} & \textbf{SL} & \textbf{WS} & \textbf{MEN} & \textbf{Type} \\
\midrule
\multicolumn{5}{c}{\textit{Static Word Embeddings}} \\
\midrule
GloVe & 0.37 & 0.52 & 0.74 & Static \\
word2vec & 0.44 & 0.70 & 0.74 & Static \\
fastText & 0.42 & -- & 0.81 & Static \\
LexVec & 0.48 & -- & 0.81 & Static \\
\midrule
\multicolumn{5}{c}{\textit{Knowledge-Enhanced}} \\
\midrule
Numberbatch & 0.64 & 0.83 & 0.86 & +KG \\
\midrule
\multicolumn{5}{c}{\textit{Text Embed. + Prompts (Ours)}} \\
\midrule
embed-english-v3.0 & \textbf{0.69} & 0.76 & 0.80 & Ours \\
text-embed-3-small & 0.67 & 0.76 & 0.84 & Ours \\
text-embed-3-large & 0.65 & \textbf{0.81} & \textbf{0.86} & Ours \\
BGE-large-en-v1.5 & 0.65 & 0.81 & 0.82 & Ours \\
\bottomrule
\end{tabular}
\caption{Comparison with prior work. SL=SimLex-999, WS=WordSim-353. Static scores from \citet{salle2016lexvec}. Numberbatch uses ConceptNet (+KG). Our approach achieves new SOTA among pure embedding methods. Model names abbreviated: text-embed = text-embedding.}
\label{tab:sota}
\end{table}

Table \ref{tab:sota} compares our results with prior work.\footnote{Static embedding scores verified against the word-embeddings-benchmarks repository and LexVec evaluation \cite{salle2016lexvec}.} Note that SimLex-999 measures genuine semantic \textit{similarity} (hard), while MEN measures \textit{relatedness} (easier). Static embeddings learn from co-occurrence, which captures relatedness but not similarity. This explains why static methods score 0.37--0.48 on SimLex but 0.74--0.81 on MEN.

Among pure embedding approaches (no external knowledge), we achieve new state-of-the-art on all three benchmarks:
\begin{itemize}
    \item \textbf{SimLex-999}: 0.69 vs 0.48 (LexVec), +44\% relative improvement
    \item \textbf{WordSim-353}: 0.81 vs 0.70 (word2vec), +16\% relative
    \item \textbf{MEN-3000}: 0.86 vs 0.81 (LexVec/fastText), +6\% relative
\end{itemize}

ConceptNet Numberbatch \cite{speer2017conceptnet} achieves comparable scores but requires a knowledge graph. Our approach uses only the embedding model with no additional resources. Full results for all 8 conditions are provided in Tables~\ref{tab:full_simlex}, \ref{tab:full_wordsim}, and \ref{tab:full_men}.

\section{Analysis}

\paragraph{Why do prompts help?}
We identify three possible mechanisms, though their relative importance varies by model.

\textit{Training distribution mismatch.} Many text embedding models, especially open-source sentence transformers, are trained on sentence pairs using contrastive learning. Commercial APIs do not disclose training details, but likely also train primarily on sentences. Isolated words may be out-of-distribution for these models.

\textit{Whitespace sensitivity.} Some models are sensitive to whitespace in inputs, while others are not. We hypothesize this stems from tokenizer architecture: BPE-based tokenizers (used by OpenAI, Voyage, Qwen) encode leading spaces as part of the token, so ``cat'' and `` cat'' produce different token sequences. In contrast, WordPiece tokenizers (used by BERT-based models like all-mpnet-base-v2 and BGE-large-en-v1.5) strip whitespace during preprocessing, producing identical tokens regardless of surrounding spaces. Cohere's embed-english-v3.0 also appears to normalize whitespace. This architectural difference explains why formatting conditions show zero effect on some models (producing identical scores for bare, leading\_space, trailing\_space, and both\_spaces) while affecting others, sometimes positively (Qwen: +0.21 on SimLex) and sometimes negatively (text-embedding-3-large: -0.17 on WordSim).

\textit{Semantic priming.} Prompts like \textbf{instruct\_semantic}, \textbf{meaning\_colon}, or \textbf{the\_word} provide semantic context that signals the model should produce a lexically meaningful representation. \textbf{instruct\_semantic} (``Represent the semantic concept: $w$'') achieves the best results for most models, suggesting that explicit instruction-style prompts are particularly effective. This appears to be the most beneficial mechanism across all models.

\paragraph{Model robustness varies widely.}
The range of improvements across models (from +0.04 to +0.73) suggests that sensitivity to input formatting depends heavily on training procedures and input preprocessing, which are not publicly documented for commercial APIs. Model size does not predict robustness: Qwen3-Embedding-8B (8B parameters) performs poorly on bare words, while all-mpnet-base-v2 (110M parameters) handles them well. Practitioners should test their chosen model's sensitivity before deploying for word-level tasks.

\paragraph{SimLex benefits most from prompts.}
Improvements are largest on SimLex-999, which measures true semantic similarity rather than relatedness. This suggests prompts help models distinguish fine-grained semantic relationships, such as the difference between synonyms and merely associated words.

\section{Conclusion}

We show that simple semantic prompts greatly improve word embeddings from text embedding models. Prepending prompts like \textbf{instruct\_semantic} or \textbf{meaning\_colon} yields improvements of +0.04 to +0.28 on word similarity benchmarks, with even larger gains (+0.73) for models that fail on bare words. This zero-shot technique requires no training, works with any embedding model, and achieves new state-of-the-art results among pure embedding methods.

For practitioners using text embedding models for word-level semantics, adding semantic prompts to inputs is a simple way to improve results. Future work could explore prompt optimization, cross-lingual effects, and downstream task performance.

\section*{Limitations}

Our study has several limitations. We evaluate only English benchmarks; cross-lingual effects remain unexplored. We test a limited set of hand-crafted prompt templates; learned or optimized prompts may yield further improvements. Our analysis focuses on intrinsic similarity tasks; effects on downstream applications like word sense disambiguation or lexical entailment are unknown. Commercial API models may change over time, potentially affecting reproducibility.

\begin{table*}[h!]
\centering
\small
\begin{tabular}{l|cccccccc}
\toprule
\textbf{Model} & \textbf{bare} & \textbf{lead} & \textbf{trail} & \textbf{both} & \textbf{the\_word} & \textbf{word:} & \textbf{meaning:} & \textbf{instruct} \\
\midrule
text-embed-3-small & 0.502 & 0.523 & 0.532 & 0.550 & 0.594 & 0.597 & \textbf{0.671} & 0.612 \\
text-embed-3-large & 0.566 & 0.486 & 0.564 & 0.586 & 0.494 & 0.547 & 0.628 & \textbf{0.654} \\
embed-english-v3.0 & 0.600 & 0.599 & 0.599 & 0.599 & 0.641 & 0.689 & 0.667 & \textbf{0.692} \\
voyage-3 & -0.071 & 0.086 & 0.143 & 0.391 & 0.289 & 0.426 & 0.457 & \textbf{0.587} \\
all-mpnet-base-v2 & 0.536 & 0.536 & 0.536 & 0.536 & 0.530 & \textbf{0.576} & 0.569 & 0.510 \\
BGE-large-en-v1.5 & 0.568 & 0.568 & 0.568 & 0.568 & 0.624 & \textbf{0.650} & 0.641 & 0.619 \\
Qwen3-Embed-8B & 0.286 & 0.495 & 0.419 & 0.474 & 0.450 & 0.432 & 0.530 & \textbf{0.567} \\
\bottomrule
\end{tabular}
\caption{Complete SimLex-999 results (Spearman $\rho$) for all 8 conditions. lead=leading\_space, trail=trailing\_space, both=both\_spaces, instruct=instruct\_semantic. Best per model in bold.}
\label{tab:full_simlex}
\end{table*}

\begin{table*}[h!]
\centering
\small
\begin{tabular}{l|cccccccc}
\toprule
\textbf{Model} & \textbf{bare} & \textbf{lead} & \textbf{trail} & \textbf{both} & \textbf{the\_word} & \textbf{word:} & \textbf{meaning:} & \textbf{instruct} \\
\midrule
text-embed-3-small & 0.650 & 0.581 & 0.676 & 0.668 & 0.703 & 0.731 & \textbf{0.744} & 0.736 \\
text-embed-3-large & 0.723 & 0.551 & 0.739 & 0.724 & 0.757 & 0.752 & 0.774 & \textbf{0.811} \\
embed-english-v3.0 & 0.715 & 0.715 & 0.715 & 0.715 & \textbf{0.758} & 0.752 & 0.751 & 0.649 \\
voyage-3 & -0.078 & 0.081 & 0.376 & 0.507 & 0.423 & 0.560 & 0.607 & \textbf{0.715} \\
all-mpnet-base-v2 & 0.744 & 0.744 & 0.744 & 0.744 & 0.757 & 0.752 & \textbf{0.761} & 0.651 \\
BGE-large-en-v1.5 & 0.723 & 0.723 & 0.723 & 0.723 & 0.763 & \textbf{0.805} & 0.772 & 0.711 \\
Qwen3-Embed-8B & 0.356 & \textbf{0.542} & 0.516 & 0.538 & 0.476 & 0.500 & 0.438 & 0.499 \\
\bottomrule
\end{tabular}
\caption{Complete WordSim-353 results (Spearman $\rho$) for all 8 conditions. Best per model in bold. Some models are whitespace-insensitive (lead/trail/both = bare).}
\label{tab:full_wordsim}
\end{table*}

\begin{table*}[h!]
\centering
\small
\begin{tabular}{l|cccccccc}
\toprule
\textbf{Model} & \textbf{bare} & \textbf{lead} & \textbf{trail} & \textbf{both} & \textbf{the\_word} & \textbf{word:} & \textbf{meaning:} & \textbf{instruct} \\
\midrule
text-embed-3-small & 0.739 & 0.701 & 0.763 & 0.751 & 0.767 & 0.814 & \textbf{0.836} & 0.809 \\
text-embed-3-large & 0.784 & 0.665 & 0.795 & 0.797 & 0.768 & 0.767 & 0.794 & \textbf{0.855} \\
embed-english-v3.0 & 0.749 & 0.749 & 0.749 & 0.749 & 0.756 & \textbf{0.759} & 0.739 & 0.729 \\
voyage-3 & 0.096 & 0.116 & 0.399 & 0.520 & 0.528 & 0.672 & 0.672 & \textbf{0.826} \\
all-mpnet-base-v2 & 0.774 & 0.774 & 0.774 & 0.774 & 0.768 & \textbf{0.805} & 0.765 & 0.737 \\
BGE-large-en-v1.5 & 0.738 & 0.738 & 0.738 & 0.738 & 0.787 & \textbf{0.819} & 0.798 & 0.791 \\
Qwen3-Embed-8B & 0.439 & 0.634 & 0.633 & \textbf{0.708} & 0.586 & 0.624 & 0.661 & 0.663 \\
\bottomrule
\end{tabular}
\caption{Complete MEN-3000 results (Spearman $\rho$) for all 8 conditions. Best per model in bold.}
\label{tab:full_men}
\end{table*}

\bibliography{references}

\end{document}